\documentclass[11pt,a4paper]{article}
\usepackage[acceptedWithA]{tacl2018v2}

\usepackage{times}
\usepackage{latexsym}
\usepackage[tone]{tipa}

\usepackage{url}
\usepackage{microtype}
\usepackage[font=small]{caption}
\usepackage{xspace}
\usepackage{adjustbox}
\usepackage{booktabs}

\usepackage{amsmath}
\usepackage{amsfonts}
\usepackage{amsthm}

\usepackage{subcaption}

\usepackage{graphicx}
\usepackage[font=small]{caption}
\usepackage{microtype}
\usepackage{mathtools}

\usepackage{color}
\usepackage{xcolor}
\definecolor{darkgrey}{rgb}{0.2,0.2,0.2}
\definecolor{grey}{rgb}{0.9,0.9,0.9}
\definecolor{darkblue}{rgb}{0.0,0.0,0.5}
\definecolor{darkred}{rgb}{0.5,0.0,0.0}
\definecolor{darkorange}{rgb}{1.0,0.55,0.0}
\definecolor{darkgreen}{rgb}{0.0,0.6,0.0}
\definecolor{darkyellow}{rgb}{1.0,0.65,0.0}
\definecolor{darkorange}{rgb}{1.0,0.65,0.0}
\definecolor{darkergreen}{rgb}{0.0,0.4,0.0}
\definecolor{lightblue}{rgb}{0.8,0.8,1.0}
\definecolor{lightgreen}{rgb}{0.8,1.0,0.8}
\definecolor{lightred}{rgb}{1.0,0.8,0.8}
\definecolor{lightyellow}{rgb}{1.0,1.0,0.8}
\definecolor{lightorange}{rgb}{1.0,0.9,0.8}
\definecolor{lightgrey}{rgb}{0.96,0.97,0.98}
\definecolor{brilliantlavender}{rgb}{0.96, 0.73, 1.0}

\definecolor{ryanred}{rgb}{0.64, 0.0, 0.0}
\definecolor{ryanblue}{rgb}{0.13, 0.0, 0.58}
\definecolor{ryangreen}{rgb}{0.12, 0.59, 0.0}
\definecolor{ryanpurple}{rgb}{0.65, 0.0, 0.57}

\definecolor{mylavender}{HTML}{BD71E1}
\definecolor{darkpurple}{HTML}{531B93}

\usepackage{xcolor}
\usepackage{color}
\usepackage{inconsolata} 
\usepackage{hyperref} 
\definecolor{linkpurple}{HTML}{45107A} 
\hypersetup{colorlinks=true, linkcolor=linkpurple, citecolor=linkpurple, urlcolor=linkpurple, filecolor=linkpurple}
\usepackage{etoolbox}
\makeatletter
\patchcmd{\@combinedblfloats}{\box\@outputbox}{\unvbox\@outputbox}{}{}
\makeatother
\apptocmd{\thebibliography}{\interlinepenalty=10000\relax}{}{}
\newcommand{\Note}[3]{\textcolor{#2}{\textbf{[#1: #3]}}}
\renewcommand{\Note}[3]{}

\usepackage{enumerate}
\usepackage[inline,shortlabels]{enumitem}

\newcommand{\word}[1]{\textit{#1}}
\newcommand{\defn}[1]{\textbf{#1}} 
\newcommand{\mydots}{%
  \ifmmode
    .\mkern-1mu.\mkern-1mu.%
  \else
    .\kern-0.1em.\kern-0.1em.%
  \fi
}
\newcommand{\mycdots}{%
  \ifmmode
    {\cdot}\mkern-1mu{\cdot}\mkern-1mu{\cdot}%
  \else
    \cdot\kern-0.1em\cdot\kern-0.1em\cdot%
  \fi
}
\renewcommand{\ldots}{\mydots}

\renewcommand{\cdots}{\mycdots}
\providecommand{\eos}{\textsc{eos}} 
\providecommand{\citepos}[1]{\citeauthor{#1}'s~\citeyearpar{#1}} 

\usepackage{cleveref}
\crefname{section}{\S}{\S\S}
\Crefname{section}{\S}{\S\S}
\crefname{table}{Table}{}
\crefname{figure}{Figure}{}
\crefname{algorithm}{Alg.}{}
\crefname{equation}{Equation}{}
\crefname{appendix}{App.}{}
\crefformat{section}{\S#2#1#3}

\newcommand{\saveForCR}[1]{}

\newcommand{\vb}{{\mathbf b}}
\newcommand{\vf}{{\mathbf f}}
\newcommand{\vphi}{{\boldsymbol \phi}} 
\newcommand{\vpi}{{\boldsymbol \pi}}   

\newcommand{\xx}{\boldsymbol{x}}
\newcommand{\yy}{\boldsymbol{y}}

\title{Recurrent Neural Networks in Linguistic Theory: Revisiting \citet{pinker1988language} and the Past Tense Debate}

\author{Christo Kirov \\
Johns Hopkins University \\
Baltimore, MD \\
{\tt \href{mailto:ckirov1@jhu.edu}{\texttt{ckirov1@jhu.edu}}} \\\And
Ryan Cotterell \\
Johns Hopkins University \\
Baltimore, MD \\
 {\tt \href{mailto:ryan.cotterell@jhu.edu}{\texttt{ryan.cotterell@jhu.edu}}} \\}

\date{}

\begin{document}
\maketitle

\begin{abstract}

Can advances in NLP help advance cognitive modeling? We examine the
role of artificial neural networks, the current state of the art in
many common NLP tasks, by returning to a classic case study.  In 1986, Rumelhart and McClelland famously introduced
a neural architecture that learned to transduce English verb
stems to their past tense forms.
Shortly thereafter,
\citet{pinker1988language} presented a comprehensive rebuttal of many
of \citepos{rumelhart1986learning} claims.
Much of the force of their attack centered on the
empirical inadequacy of the \citet{rumelhart1986learning} model.
Today, however, that model is severely
outmoded. We show that the encoder--decoder network architectures used in
modern NLP systems obviate most of \citepos{pinker1988language} criticisms without requiring any simplification of the past tense mapping problem.  We
suggest that the empirical performance of modern networks warrants a re{\"e}xamination of their utility in
linguistic and cognitive modeling.

\end{abstract}

\section{Introduction}
In their famous 1986 opus, \citet{rumelhart1986learning} describe a
neural network capable of transducing English verb stems to their past
tense. The strong cognitive claims in the article fomented
a veritable brouhaha in the linguistics community that
eventually led to the highly influential rebuttal of
\citet{pinker1988language}. \citeauthor{pinker1988language} highlighted the extremely poor
empirical performance of the \citeauthor{rumelhart1986learning} model, and pointed out a number of
theoretical issues with the model, that they suggested would apply to \emph{any}
neural network, contemporarily branded connectionist approaches. Their critique was so successful that many
linguists and cognitive scientists to this day do not consider neural networks a viable approach to modeling linguistic data and human cognition.\looseness=-1

In the field of natural language processing (NLP), however, neural networks have experienced a renaissance. With novel
architectures, large new datasets available for training, and access
to extensive computational resources, neural networks now constitute the
state of the art in many NLP tasks. However, NLP as a discipline has a
distinct practical bent and more often concerns itself with the large-scale
engineering applications of language technologies. As such,
the field's findings are not always considered relevant to the
scientific study of language, i.e., the field of linguistics.
Recent work, however, has indicated that this perception is changing, with researchers, for example, probing the ability of neural networks to learn syntactic dependencies like subject--verb agreement \cite{linzen2016assessing}.\looseness=-1

Moreover, in the domains of morphology and
phonology, both NLP practitioners and linguists have considered
virtually identical problems, seemingly unbeknownst to each other. For example, both computational and
theoretical morphologists are concerned with how different inflected
forms in the lexicon are related and how one can learn to generate such inflections from data. Indeed, the original \citeauthor{rumelhart1986learning} network focuses on such a generation task,
i.e., generating English past tense forms from their stems. \citepos{rumelhart1986learning} network, however, was severely limited and did not
generalize correctly to held-out data. In contrast, state-of-the-art
morphological generation networks used in NLP, built from the modern evolution of Recurrent Neural Networks (RNNs) explored by \citet{elman1990finding} and others,
 solve the same
problem almost perfectly \cite{cotterell-et-al-2016-shared}. This level of performance on a cognitively
relevant problem suggests that it is time to consider further incorporating network modeling into the study of linguistics and cognitive science.\looseness=-1

Crucially, we wish to sidestep one of the issues that framed the
original debate between \citeauthor{pinker1988language} and \citeauthor{rumelhart1986learning}---whether or not neural models learn and
use `rules'.  From our perspective, any system that picks up
systematic, predictable patterns in data may be referred
to as rule-governed. We focus instead on an empirical assessment of the
ability of a modern state-of-the-art neural architecture to learn 
linguistic patterns, asking the following questions:
\begin{enumerate*}[(i)]
\item Does the learner induce the full set of correct generalizations
  about the data? Given a range of novel inputs, to what extent does
  it apply the correct transformations to them?
\item Does the behavior of the learner mimic humans? Are the errors human-like?
\end{enumerate*}

In this work, we run new experiments examining the
ability of the encoder--decoder architecture developed for
machine translation
\cite{DBLP:conf/nips/SutskeverVL14,DBLP:journals/corr/BahdanauCB14} to
learn the English past tense. The results suggest that modern nets absolutely meet
the first criterion above, and often meet the second. Furthermore,
they do this given limited prior knowledge of linguistic
structure: the networks we consider do not have phonological features
built into them and must instead learn their own representations for
input phonemes. The design and performance of these networks invalidate
many of the criticisms in \citet{pinker1988language}. 
We contend that, given the gains displayed in this case study, which is characteristic of problems in the morpho-phonological domain, researchers across linguistics and cognitive science should consider evaluating modern neural architectures as part of their modeling toolbox.

This paper is structured as follows. \Cref{sec:english-past-tense} describes the problem under consideration, the English past tense. \Cref{sec:from-rm-to-ed} lays out the original Rumelhart and McClelland model from 1986 in modern machine-learning parlance and compares it to a state-of-the-art encoder--decoder architecture. A historical perspective on alternative approaches to modeling, both neural and non-neural, is provided in \cref{sec:related-work}. The empirical performance of the encoder--decoder architecture is evaluated in \cref{sec:evaluation}. \Cref{sec:crit-summary} provides a summary of which of \citepos{pinker1988language} original criticisms have effectively been resolved, and which ones still require further consideration. Concluding remarks follow.\looseness=-1

\section{The English Past Tense}\label{sec:english-past-tense}

\begin{table}
  \begin{adjustbox}{width=1.\columnwidth}
    \begin{tabular}{lll lll l} \toprule
      \multicolumn{3}{c}{orthographic} & \multicolumn{3}{c}{IPA} & \\ \cmidrule(lr){1-3} \cmidrule(lr){4-6} \cmidrule(lr){7-7}
    stem & past & part. & stem & past & part. & \multicolumn{1}{c}{infl. type} \\ \midrule
    go & went & gone & go\textipa{U} & w\textipa{E}nt & go\textipa{U}n & suppletive \\
    sing & sang    & sung &  s\textipa{I}\textipa{N} & s\textipa{\ae}\textipa{N}  & s\textipa{U}\textipa{N} & ablaut  \\
    swim & swam    & swum &  sw\textipa{I}m & sw\textipa{\ae}m & sw\textipa{U}m  & ablaut \\
    sack & sacked  & sacked & s\textipa{\ae}k & s\textipa{\ae}kt & s\textipa{\ae}kt & [-t]\\
    sag  & sagged  & sagged & s\textipa{\ae}g & s\textipa{\ae}gd & s\textipa{\ae}gd & [-d] \\
    pat  & patted  & patted & p\textipa{\ae}t & p\textipa{\ae}t\textipa{I}{d} & p\textipa{\ae}t\textipa{I}{d} & [-\textipa{I}d] \\
    pad  & padded  & padded & p\textipa{\ae}d & p\textipa{\ae}d\textipa{I}{d} & p\textipa{\ae}d\textipa{I}{d} & [-\textipa{I}d] \\
    \bottomrule
  \end{tabular}
 \end{adjustbox}
  \caption{Examples of inflected English verbs.}
    \label{tab:english-verbs}
\end{table}

Many languages mark words with syntactico-semantic distinctions. For instance, English marks the distinction between
present and past tense verbs, e.g., \word{walk} and
\word{walked}. English verbal morphology is relatively impoverished,
distinguishing maximally five forms for the copula \word{to be} and
only three forms for most verbs. In this work, we consider learning to conjugate
the English verb forms, rendered as phonological strings. As it is the focus of \citepos{rumelhart1986learning} original study, we focus primarily on the English past tense
formation.\looseness=-1

Both regular and irregular patterning exist in English.
Orthographically, the canonical regular suffix is \word{-ed}, which,
phonologically, may be rendered as one of three phonological strings:
[-\textipa{I}d], [-d] or [-t]. The choice among the three is
deterministic, depending only on the phonological properties of the
previous segment. English selects [-\textipa{I}d] where the previous
phoneme is a [t] or [d],
e.g. [p\textipa{\ae}t]$\mapsto$[p\textipa{\ae}t\textipa{I}d] (\word{pat}$\mapsto$\word{patted}) and [p\textipa{\ae}d]$\mapsto$[p\textipa{\ae}d\textipa{I}d] (\word{pad}$\mapsto$\word{padded}). In other
cases, English enforces voicing agreement: it opts for [-d] when the
preceding phoneme is a voiced consonant or a vowel, e.g.,
[s\textipa{\ae}g]$\mapsto$[s\textipa{\ae}gd] (\word{sag}$\mapsto$\word{sagged})
and [\textipa{S}o\textipa{U}]$\mapsto$[\textipa{S}o\textipa{U}d] (\word{show}$\mapsto$\word{showed}), and for [-t] when the preceding phoneme
is an unvoiced consonant, e.g., [s\textipa{\ae}k]$\mapsto$[s\textipa{\ae}kt]
(\word{sack}$\mapsto$\word{sacked}).
English irregulars are either suppletive, e.g.,
[go\textipa{U}]$\mapsto$[w\textipa{E}nt] (\word{go}$\mapsto$\word{went}), or exist in sub-regular islands defined by processes like ablaut, e.g., \word{sing}$\mapsto$\word{sang}, that may contain several verbs \cite{nelson2010english}: 
see \cref{tab:english-verbs}.

\paragraph{Single versus Dual Route.}
A frequently discussed cognitive aspect of past tense processing concerns whether or not irregular forms have their own processing pipeline in the brain. \citet{pinker1988language} proposed separate modules for regular and irregular verbs; regular verbs go through a general, rule-governed
transduction mechanism, and exceptional irregulars are produced via simple memory-lookup.\footnote{Note that irregular lookup can simply be recast as the application of a context-specific rule.} While some studies, e.g., \cite{marslen1997dissociating,ullman1997neural}, provide corroborating evidence from speakers with selective impairments to regular or irregular verb production, others have called these results
into doubt \cite{stockall2006single}. From the perspective of this paper, a complete model of the English past tense should cover both regular and irregular transformations. The neural network approaches we advocate for achieve this goal, but do not clearly fall into either the single or dual-route category---internal computations performed by each network remain opaque, so we cannot at present make a claim whether two separable computation paths are present.

\subsection{Acquisition of the Past Tense}
The English past tense is of considerable
theoretical interest due to the now well-studied acquisition patterns of
children. As first shown by \citet{berko1958child} in the so-called
wug-test, knowledge of English morphology {\em cannot} be attributed solely
to memorization. Indeed, both adults and children are fully capable of
generalizing the patterns to novel words, e.g., [w\textipa{2}g]$\mapsto$[w\textipa{2}gd] (\word{wug}$\mapsto$\word{wugged}).
During acquisition, only a few types of errors are common; children rarely blend regular and
irregular forms, e.g., the past tense of \word{come} is either produced
as \word{comed} or \word{came}, but rarely \word{camed} \cite{pinker2015words}.

\paragraph{Acquisition Patterns for Irregular Verbs.}
It is widely claimed that children learning the past tense forms of irregular verbs exhibit a `U-shaped' learning curve.
At first, they correctly conjugate irregular
forms, e.g., \word{come}$\mapsto$\word{came}, then they regress during a
period of \defn{overregularization} producing the past tense as \word{comed} as they acquire the general past tense formation.  Finally,
they learn to produce both the regular and irregular
forms. Plunkett \& Marchman, however, observed a more nuanced form of this behavior. Rather than a \emph{macro} U-shaped learning process that applies globally and uniformly to all irregulars, they noted that many irregulars \emph{oscillate} between correct and overregularized productions \cite{marchman1988rules}. These oscillations, which Plunkett \& Marchman refer to as a \emph{micro} U-shape, further apply at different rates for different verbs \cite{plunkett1991u}.
Interestingly, while the exact pattern of irregular acquisition may be disputed, children rarely \defn{overirregularize},
i.e., misconjugate a regular verb as if it were irregular such as 
\word{ping}$\mapsto$\word{pang}. 

\section{1986 vs Today}\label{sec:from-rm-to-ed}

In this section, we compare the original \citeauthor{rumelhart1986learning} architecture from 1986, to today's state-of-the-art neural architecture for morphological transduction, the encoder--decoder model.

\subsection{\citet{rumelhart1986learning}}\label{sec:rm}
For many linguists, the face of neural networks to this day
remains the work of \citet{rumelhart1986learning}.  Here, we describe in detail their original architecture, using modern machine learning parlance whenever
possible. Fundamentally, \citet{rumelhart1986learning} were interested in designing a
sequence-to-sequence network for variable-length input using a
small feed-forward network. From an NLP perspective, this
work constitutes one of the first attempts to design a network for a
task reminiscent of popular NLP tasks today that require
variable-length input, e.g., part-of-speech tagging, parsing and generation.

\paragraph{Wickelphones and Wickelfeatures.}
Unfortunately, a fixed-sized feed-forward network is not immediately compatible with the goal of transducing sequences of varying lengths.
Rumelhart and McClelland decided to get around this limitation by 
 representing each string as the set of its constituent phoneme trigrams.
Each trigram is termed a \defn{Wickelphone}
\cite{wickelgran1969context}. As a concrete example, the IPA-string
     [\#k\textipa{\ae}t\#], marked with a special beginning- and
     end-of-string character, contains three distinct Wickelphones:
     [\#k\textipa{\ae}], [k\textipa{\ae}t],
     [\textipa{\ae}t\#]. 
     In fact, Rumelhart and McClelland went one step
     further and decomposed Wickelphones into component \defn{Wickelfeatures}, or trigrams of phonological features, one for each Wickelphone phoneme.
       For example, the Wickelphone [ipt] is represented by the
       Wickelfeatures $\langle$$+$vowel,$+$unvoiced,$+$interrupted$\rangle$ and $\langle$$+$high,$+$stop,$+$stop$\rangle$.
       Since there are far fewer Wickelfeatures than
       Wickelphones, words could be represented with fewer units
       (of key importance for 1986 hardware) and more shared Wickelfeatures
       potentially meant better generalization. 

We can describe \citepos{rumelhart1986learning} representations using the
modern linear-algebraic notation standard among researchers in neural
networks.  First, we assume that the language under consideration
contains a fixed set of phonemes $\Sigma$, plus an edge symbol \# marking the beginning and end of
words. Then, we construct the set of all Wickelphones $\Phi$ and the set
of all Wickelfeatures ${\cal F}$ by enumeration.  The first layer of the \citeauthor{rumelhart1986learning} neural network consists of two deterministic functions: (i)
$\vphi : \Sigma^* \rightarrow \mathbb{B}^{|\Phi|}$ and (ii) $\vf :
\mathbb{B}^{|\Phi|} \rightarrow \mathbb{B}^{|{\cal F}|}$, where we
define $\mathbb{B} = \{-1, 1\}$. The first function $\vphi$ maps a
phoneme string to the set of Wickelphones that fire, as it were, on that
string, e.g., $\vphi\left(\text{[\#k\textipa{\ae}t\#]}\right) = \{ \text{[\#k\textipa{\ae}], [k\textipa{\ae}t],
     [\textipa{\ae}t\#]}
\}$. The output subset of $\Phi$ may be represented by a
binary vector of length $|\Phi|$, where a $1$ means that the
Wickelphone appears in the string and a $-1$ that it does not.\footnote{We have chosen $-1$ instead of the more traditional $0$
  so that the objective function that Rumelhart and McClelland
  optimize may be more concisely written.}  The second function
$\vf$ maps a set of Wickelphones to its corresponding set of Wickelfeatures.

\paragraph{Pattern Associator Network.}
Here we define the complete network of \citeauthor{rumelhart1986learning}. 
We denote strings of phonemes
as $\xx \in \Sigma^*$, where $x_i$ is the $i^\text{th}$ phoneme in a string.
Given source and target
phoneme strings $\xx^{(m)}, \yy^{(m)} \in \Sigma^*$, \citeauthor{rumelhart1986learning} optimize the
following objective, a sum over the individual losses for each of the $m=1,\ldots,M$ training items:
\begin{equation}
  \begin{split}
    \sum_{m=1}^{M} \Big|\Big|\max\Big\{0, {}&-\vpi(\yy^{(m)}) \\
    &\odot \big(\mathbf{W}\vpi(\xx^{(m)}) + \vb\big)\Big\}\Big|\Big|_1,
  \end{split}
  \label{eq:rm}
\end{equation}
where $\max\{\cdot\}$ is taken point-wise, $\odot$ is point-wise multiplication, $\mathbf{W} \in \mathbb{R}^{|{\cal F}|\times |{\cal F}|}$ is a projection matrix, $\vb \in \mathbb{R}^{|{\cal F}|}$
is a bias term, and $\vpi = \vf \circ \vphi$ is the composition of the Wickelphone
and Wickelfeature encoding functions. Using modern terminology, the architecture is
a linear model for a multi-label classification problem \cite{tsoumakas2006multi}: the goal is to predict
the {\em set} of Wickelfeatures in the target form $\yy^{(m)}$ given the input form $\xx^{(m)}$
using a point-wise perceptron loss (hinge loss without a margin), i.e., a binary perceptron
predicts each feature independently, but there is one set of parameters \{$\mathbf{W}$, $\vb$\} . The total loss incurred
is the {\em sum} of the per-feature loss, hence the use of the $L_1$ norm.
The model is trained with stochastic sub-gradient descent (the perceptron update rule; \citealp{rosenblatt1958perceptron,bertsekas2015convex}) with a fixed learning rate.\footnote{Follow-up work, e.g., \citet{plunkett1991u}, has speculated
  that the original experiments in \citet{rumelhart1986learning} may not have converged.
  Indeed, convergence may not be guaranteed depending on the fixed learning rate chosen. As \cref{eq:rm} is jointly convex
  in its parameters $\{\mathbf{W}, \vb\}$, there exist convex optimization algorithms that will guarantee
  convergence, albeit often with a decaying learning rate.\looseness=-1
  }
  Later work augmented the architecture with multiple layers and
  nonlinearities \cite[Table 3.3]{marcus2001algebraic}. 

  \paragraph{Decoding.}
  Decoding the \citeauthor{rumelhart1986learning} network necessitates solving a tricky optimization
  problem. Given an input phoneme string $\xx^{(m)}$, we then must find the corresponding
  $\yy \in \Sigma^*$ that minimizes:
\begin{equation}
\left|\left| \vpi(\yy) - \text{threshold}\left\{\mathbf{W}\vpi(\xx^{(m)}) + \vb\right\}\right|\right|_0,
\end{equation}
where $\text{threshold}$ is a step function that maps all non-positive reals to $-1$ and all positive reals to $1$.
In other words, we seek the phoneme string $\yy$ that shares the most features
with the maximum a-posteriori (MAP) decoded binary vector. This problem is intractable, and so
\citet{rumelhart1986learning} provide an approximation. For each test stem, they hand-selected a set of likely past tense candidate forms, e.g.,
good candidates for the past tense of \word{break} would be $\{$\word{break}, \word{broke}, \word{brake}, \word{braked}$\}$, and choose the form with Wickelfeatures closest to the network's output. 
This manual approximate decoding procedure is not intended to be cognitively plausible.

\paragraph{Architectural Limitations.}
 \citeauthor{rumelhart1986learning} used Wickelphones and Wickelfeatures in order to help with
 generalization and limit their network to a tractable size. However,
 this came at a significant cost to the network's ability to represent
 unique strings---the encoding is lossy: two words may have the
 same set of Wickelphones or features.  The easiest way to see this
 shortcoming is to consider morphological reduplication, which is
 common in many of the world's languages. \citeauthor{pinker1988language} provide an example from
 the Australian language of Oykangand which distinguishes between \word{algal} `straight' and \word{algalgal} `ramrod straight'; both of
 these strings have the identical Wickelphone set $\{$[{\em \#al}], [{\em
   alg}], [{\em lga}], [{\em gal}], [{\em al\#}]$\}$.
Moreover, \citeauthor{pinker1988language} point out that phonologically related words such as
[sl\textipa{I}t] and [s\textipa{I}lt] have disjoint sets of
Wickelphones: $\{$[\#sl], [sl\textipa{I}], [l\textipa{I}t],
[\textipa{I}t\#]$\}$ and $\{$[\#s\textipa{I}], [s\textipa{I}l],
[\textipa{I}lt], [lt\#]$\}$, respectively. These two words differ only
by an instance of metathesis, or swapping the order of nearby sounds. The use of Wickelphone representations imposes the strong claim that they have nothing in common phonologically, despite sharing all phonemes. \citeauthor{pinker1988language} suggest this is unlikely to be the case. As one point of evidence, the metathesis of the kind that differentiates [sl\textipa{I}t] and [s\textipa{I}lt] is a common diachronic change. In English, for example, [horse] evolved from [hross], and [bird] from [brid] \cite{Jesperson1942}.

\subsection{Encoder--Decoder Architectures}\label{sec:rnns}
The NLP community has recently developed an analogue to the past tense
generation task originally considered by
\citet{rumelhart1986learning}: morphological paradigm completion
\cite{durrett2013supervised,NicolaiCK15,TACL480,AhlbergFH15,faruqui-kumar:2015:NAACL-HLT}. The
goal is to train a model capable of mapping the lemma (stem in the case
of English) to each form in the paradigm.  In the case of
English, the goal would be to map a lemma, e.g., \word{walk}, to its
past tense word \word{walked} as well as its gerund and third person
present singular \word{walking} and \word{walks},
respectively. This task generalizes the \citeauthor{rumelhart1986learning} setting in that it requires learning more mappings than simply lemma
to past tense.

By definition, any system that solves the more general morphological paradigm completion task must also be able to solve the original \citeauthor{rumelhart1986learning} task. As we wish to highlight the strongest currently available alternative to \citeauthor{rumelhart1986learning}, we focus on the state-of-the-art in morphological paradigm completion: the encoder--decoder network architecture
\cite{cotterell-et-al-2016-shared}.  This architecture consists of two
recurrent neural networks (RNNs) coupled together by an attention
mechanism. The encoder RNN reads each symbol in the input string one
at a time, first assigning it a unique embedding, then processing that
embedding to produce a representation of the phoneme given the rest of
the phonemes in the string. The decoder RNN produces a sequence of
output phonemes one at a time, using the attention mechanism to peek
back at the encoder states as needed. Decoding ends when the end-of-string symbol $\eos$
is output. Formally, the encoder--decoder model is autoregressive: it factors the probability of an output string $\yy \in \Sigma^*$ into per-symbol conditionals, each a softmax over the augmented alphabet $\Sigma \cup \{\eos\}$ (of size $|\Sigma|+1$),
\begin{equation}
  \vec{p}(y_n \mid \yy_{<n}, \xx) = \frac{\exp(o_{n, y_n})}{\sum_{y \in \Sigma \cup \{\eos\}} \exp(o_{n, y})},
  \label{eq:softmax}
\end{equation}
where the logit vector $\mathbf{o}_n = g(\mathbf{s}_n, \mathbf{c}_n) \in \mathbb{R}^{|\Sigma|+1}$ has one component $o_{n, y}$ per candidate symbol $y \in \Sigma \cup \{\eos\}$, produced by a non-linear function $g$ (a multi-layer perceptron) of the decoder hidden state $\mathbf{s}_n$ and the attention context $\mathbf{c}_n$ (defined below). The probability of a \emph{complete} string then factors into a prefix probability and a halting probability,
\begin{subequations}
\begin{align}
  p(\yy \mid \xx) &= \vec{p}(\yy \mid \xx)\; \vec{p}(\eos \mid \yy, \xx), \label{eq:ed-string}\\
  \vec{p}(\yy \mid \xx) &= \prod_{n=1}^{N} \vec{p}(y_n \mid \yy_{<n}, \xx), \label{eq:ed-prefix}
\end{align}
\end{subequations}
where $\vec{p}(\yy \mid \xx)$ is the \emph{prefix probability}---the probability that the model generates $\yy$ as a prefix, before emitting $\eos$---and multiplying by the halting probability $\vec{p}(\eos \mid \yy, \xx)$ gives the probability of the complete string, so that the length $N = |\yy|$ is itself generated. Here $\yy = y_1\cdots y_N$ is the output sequence, with $y_1, \ldots, y_N \in \Sigma$ and $\yy_{<n} = y_1\cdots y_{n-1}$ its length-$(n{-}1)$ prefix. Finally, $\mathbf{c}_n$ is an attention-weighted sum of the encoder hidden states $\mathbf{h}_k$, namely $\mathbf{c}_n = \sum_{k=1}^{|\xx|} \alpha_{k}(\mathbf{s}_{n-1})\, \mathbf{h}_{k}$, with attention weights $\alpha_{k}(\mathbf{s}_{n-1})$ computed from the previous decoder state.

In contrast to the \citeauthor{rumelhart1986learning} network, the encoder--decoder network
optimizes the log-likelihood of the training data, i.e., $\sum_{m=1}^M \log p(\yy^{(m)} \mid \xx^{(m)})$ for $m=1,\ldots,M$ training items.
We refer the reader to \citet{DBLP:journals/corr/BahdanauCB14} for the complete architectural specification of the specific encoder--decoder model we apply in this paper.\looseness=-1

\paragraph{Theoretical Improvements.}
While there are a number of possible architectural variants of the encoder--decoder architecture \cite{luong2015effective},\footnote{For the experiments in this paper, we use the variant in \cite{DBLP:journals/corr/BahdanauCB14}, which has explicitly been shown to be state-of-the-art in morphological transduction \cite{cotterell-et-al-2016-shared}.\looseness=-1} they all share several critical features that make up for many of the theoretical shortcomings of the
feed-forward \citeauthor{rumelhart1986learning} model. The encoder reads in each phoneme 
sequentially, preserving identity and order
and allowing any string of arbitrary length to receive a unique representation. Despite
this encoding, a flexible notion of string similarity is also
maintained as the encoder--decoder model learns a fixed embedding for each phoneme
that forms part of the representation of all strings that share the
phoneme. When the network encodes [s\textipa{I}lt] and
[sl\textipa{I}t], it uses the same phoneme 
embeddings---only the order changes. Finally, the decoder permits
sampling and scoring arbitrary length {\em fully formed strings} in polynomial time (forward sampling), so there
is no need to determine which string a non-unique set of
Wickelfeatures represents. However, we note that decoding the 1-best
string from a sequence-to-sequence model is likely NP-hard. (1-best string
decoding is even hard for weighted FSTs \cite{goodman1998parsing}.)

\paragraph{Multi-Task Capability.}\label{sec:multi-task}
A single encoder--decoder model is easily adapted to multi-task learning
\cite{caruana1997multitask,collobert2011natural}, where each task is a
single transduction, e.g., stem$\mapsto$past. Note \citeauthor{rumelhart1986learning} would need
a separate network for each transduction, e.g., stem$\mapsto$gerund and
stem$\mapsto$past participle. In fact, the current state of the art in NLP is to learn all morphological transductions in a paradigm jointly. The
key insight is to construct a single network $p(\yy \mid
\xx, t)$ to predict all inflections, marking the transformation in the input string, i.e., we feed the network
   the string  ``{\small {\tt w a l k} {\tt <gerund>}}'' as input, augmenting
   the alphabet $\Sigma$ to include morphological descriptors. We refer the reader to \citet{kann-schutze:2016:SIGMORPHON} for the
   encoding details. Thus, one network predicts all forms, e.g., $p(\yy\! \mid\! \xx\!\!=\!\!
\textit{walk}, t \!\!=\!\! \text{\small \sf past})$ yields a distribution over past
tense forms for \word{walk} and $p(\yy \!\mid\! \xx\!\!=\!\!\textit{walk}, t
\!\!=\!\! \text{\small \sf gerund})$ yields a distribution over gerunds.\looseness=-1

\section{Related Work}\label{sec:related-work}

In this section, we first describe direct followups to the original 1986 \citeauthor{rumelhart1986learning} model, employing various neural architectures. Then we review competing non-neural systems of context-sensitive rewrite rules
in the style of the Sound Pattern of English (SPE; \citealp{halle1968sound}), as favored by Pinker and
Prince.

\subsection{Further Work Over the Years}
Following \citeauthor{rumelhart1986learning}, a cottage industry devoted to cognitively plausible connectionist models of inflection learning sprouted in the linguistics and cognitive science literature. 
We provide a summary listing of the various proposals, along with broadbrush comparisons, in \cref{tab:related-work}.
While many of the approaches apply more modern feed-forward architectures than \citeauthor{rumelhart1986learning}, introducing multiple layers connected by nonlinear transformations, most continue to use feed-forward architectures with limited ability to deal with variable-length inputs and outputs, and remain unable to produce and assign probability to arbitrary output strings.

\citet{macwhinney1991implementations,plunkett1991u,plunkett1993rote,plunkett1999connectionist} map phonological strings to phonological strings using feed-forward networks, but rather than turning to Wickelphones to imprecisely represent strings of any length, they use fixed-size input and output templates, with units representing each possible symbol at each input and output position. For example, \citet{plunkett1991u,plunkett1993rote} simplify the past-tense mapping problem by only considering a language of artificially generated words of exactly three syllables and a limited set of constructed past-tense formation patterns. \citet{macwhinney1991implementations,plunkett1999connectionist} additionally modify the input template to include extra units marking particular transformations (e.g., \emph{past} or \emph{gerund}), enabling their network to learn multiple mappings.

Some proposals simplify the problem even further, mapping fixed-size inputs into a small finite set of categories, solving a classification problem rather than a transduction problem. \citet{hahn2000german,nakisa1996defaults} classify German noun stems into their appropriate plural inflection classes. \citet{plunkett1997connectionist} do the same for Arabic stems.
\citet{hoeffner1992rules}, \citet{hare1995learning}, and \citet{cottrell1994acquiring} also solve an alternative problem---mapping semantic representations (usually one-hot vectors with one unit per possible word type, and one unit per possible inflection) to phonological outputs. As these networks use unstructured semantic inputs to represent words, they must act as memories---the phonological content of any word must be memorized. This precludes generalization to word types that were not seen during training. 
Of the proposals that map semantics to phonology, the architecture in \cite{hoeffner1992rules} is unique in that it uses an attractor network rather than a feed-forward network, with the main difference being training using Hebbian learning rather than the standard backpropagation algorithm. \citet{cottrell1994acquiring} present an early use of a simple recurrent network \cite{elman1990finding} to decode output strings, making their model capable of variable length output.

\citet{bullinaria1997modeling} is one of the few models proposed that can deal with variable length \emph{inputs}. They use a derivative of the NETtalk pronunciation model \cite{sejnowski1987parallel} that would today be considered a convolutional network. Each input phoneme in a stem is read independently along with its left and right context phonemes within a limited context window (i.e., a convolutional kernel). Each kernel is then mapped to zero or more output phonemes within a fixed template. Since each output fragment is independently generated, the architecture is limited to learning only \emph{local} constraints on output string structure. Similarly, the use of a fixed context window also means that inflectional patterns that depend on long-distance dependencies between input phonemes cannot be captured.

Finally, the model of \cite{westermann1995connectionist} is arguably the most similar to a modern encoder--decoder architecture, relying on simple recurrent networks to both encode input strings and decode output strings. However, the model was intended to explicitly instantiate a dual route mechanism and contains an additional explicit memory component to memorize irregulars. Despite the addition of this memory, the model was unable to fully learn the mapping from German verb stems to their participle forms, failing to capture the correct form for strong training verbs including the copular \word{sein}$\mapsto$\word{gewesen}. As the authors note, this may be due to the difficulty of training simple recurrent networks, which tend to converge to poor local minima. Modern RNN varieties, such as the LSTMs in the encoder--decoder model, were specifically designed to overcome these training limitations \cite{hochreiter1997long}.

\begin{table*}
      \centering
  \begin{adjustbox}{width=2.0\columnwidth}
  \begin{tabular}{llll} \toprule
    Type of Model & Reference & Input & Output \\ \midrule
    Feedforward Network & \citet{rumelhart1986learning} & Wickelphones & Wickelphones \\
    Feedforward Network & \citet{macwhinney1991implementations} & Fixed Size Phonological Template & Fixed Size Phonological Template \\
    Feedforward Network & \citet{plunkett1991u} & Fixed Size Phonological Template & Fixed Size Phonological Template \\
    Attractor Network & \citet{hoeffner1992rules} & Semantics & Fixed Size Phonological Template\\
    Feedforward Network & \citet{plunkett1993rote} & Fixed Size Phonological Template & Fixed Size Phonological Template \\
    Recurrent Neural Network &  \citet{cottrell1994acquiring} & Semantics & Phonological String \\
    Feedforward Network & \citet{hare1995default} & Fixed Size Phonological Template & Inflection Class\\
    Feedforward Neural Network & \citet{hare1995learning} & Semantics & Fixed Size Phonological Template  \\
    Recurrent Neural Network & \citet{westermann1995connectionist} & Phonological String & Phonological String \\
    Feedforward Neural Network & \citet{nakisa1996defaults} & Fixed Size Phonological Template & Inflection Class\\
    Convolutional Neural Network & \citet{bullinaria1997modeling} & Phonological String & Phonological String \\
    Feedforward Neural Network & \citet{plunkett1997connectionist} & Fixed Size Phonological Template & Inflection Class \\
    Feedforward Neural Network & \citet{plunkett1999connectionist} & Fixed Size Phonological Template & Fixed Size Phonological Template \\
    Feedforward Neural Network & \citet{hahn2000german} & Fixed Size Phonological Template & Inflection Class \\
    \bottomrule
  \end{tabular}
  \end{adjustbox}
  \caption{A curated list of related work, categorized by aspects of the technique. Based on a similar list found in \citet[page 82]{marcus2001algebraic}.
  }
    \label{tab:related-work}
\end{table*}

\subsection{Non-neural Learners}

\citeauthor{pinker1988language} describe several basic ideas that underlie a traditional,
symbolic, rule-learner. Such a learner produces SPE-style rewrite
rules that may be applied to deterministically transform the input
string into the target. Rule candidates are found by comparing the stem 
and the inflected form, treating the portion that changes as the rule 
which governs the transformation. This is typically a set of non-copy edit
operations. 
If multiple stem-past pairs share similar changes, these
may be collapsed into a single rule by generalizing over the shared
phonological features involved in the changes. For example, if
multiple stems are converted to the past tense by the addition of the
suffix [-d], the learner may create a collapsed rule that adds the
suffix to all stems that end in a [+voice] sound. Different rules may
be assigned weights (e.g., probabilities) derived from how many stem-past pairs exemplify the rules.
These weights decide which rules to apply to produce the
past tense.

Several systems that follow the rule-based template above have been developed in NLP. While the SPE itself does not impose
detailed restrictions on rule structure, these systems
generate rules that can be compiled into finite-state transducers
\cite{kaplan1994regular,AhlbergFH15}. While these
systems generalize well, 
even the most successful
variants have been shown to perform significantly worse than
state-of-the-art neural networks at morphological
inflection, often with a $>$10\% differential in accuracy on
held-out data \cite{cotterell-et-al-2016-shared}.\looseness=-1

In the linguistics literature, the most straight-forward, direct machine-implemented instantiation
of the \citeauthor{pinker1988language} proposal is, arguably, the Minimal Generalization Learner
(MGL) of \citet{Albright:03} (cf.\ \citet{allen2015learning}; \citet{taatgen2002children}). This model
takes a mapping of phonemes to phonological features and makes
feature-level generalizations like the post-voice [-d] rule described
above. For a detailed technical description, see
\citet{albright2002modeling}. We treat the MGL as a baseline in  \cref{sec:evaluation}.

Unlike \citet{taatgen2002children}, which explicitly accounts for dual route processing by including both memory retrieval and rule application submodules, \citet{Albright:03} and \citet{allen2015learning} rely on discovering and correctly weighting (using weights learned by log-linear regression) highly stem-specific rules to account for irregular transformations.

Within the context of rule-based systems, several proposals focus on the question of rule generalization, rather than rule synthesis. That is, given a set of pre-defined rules, the systems implement metrics to decide whether rules should generalize to novel forms,  depending on the number of exceptions in the data set. \citet{yang2016price} defines the `tolerance principle,' a threshold for exceptionality beyond which a rule will fail to generalize. \citet{o2011productivity} treat the question of whether a rule will generalize as one of optimal Bayesian inference.

\section{Evaluation of the Encoder--Decoder}\label{sec:evaluation}

We evaluate the performance of the encoder--decoder
architecture in light of the criticisms \citeauthor{pinker1988language} levied
against the original \citeauthor{rumelhart1986learning} model. We show that in most cases, these
criticisms no longer apply.\footnote{Datasets and code for all experiments are available at \url{https://github.com/ckirov/RevisitPinkerAndPrince}}
The most potent line of attack \citeauthor{pinker1988language} use against the \citeauthor{rumelhart1986learning} model is that
it simply does not learn the English past tense very well. While the non-deterministic, manual, and non-precise decoding
procedure used by \citeauthor{rumelhart1986learning} makes it difficult to obtain exact accuracy
numbers, \citeauthor{pinker1988language} estimate that the model only prefers
the correct past tense form for about 67\% of English verb stems.
Furthermore, many of the errors made by the \citeauthor{rumelhart1986learning} network are unattested
in human performance. For example, the model produces blends of
regular and irregular past-tense formation, e.g., \word{eat}$\mapsto$\word{ated}, that children do not produce unless they mistake
\word{ate} for a present stem \cite{pinker2015words}. Furthermore, the
\citeauthor{rumelhart1986learning} model frequently produces
irregular past tense forms when a regular formation is
expected, e.g., \word{ping}$\mapsto$\word{pang}. Humans are more likely to overregularize. These
behaviors suggest the \citeauthor{rumelhart1986learning} model learns the wrong kind of
generalizations. As shown below, the encoder--decoder architecture seems to avoid these pitfalls, while outperforming a \citeauthor{pinker1988language}-style non-neural baseline.\looseness=-1

\subsection{Experiment 1: Learning the Past Tense}\label{sec:exp1}

In the first experiment, we seek to show: (i) the encoder--decoder
model successfully learns to conjugate both regular and irregular
verbs in the training data, and generalizes to held-out data at
convergence and (ii) the pattern of errors the model exhibits is compatible with attested speech errors.

\paragraph{CELEX Dataset.}
Our base dataset consists of 4039 verb types in the CELEX database
\cite{baayen1993celex}. Each verb is associated with a present
tense form (stem) and past tense form, both in IPA. Each verb is also marked as
regular or irregular \cite{Albright:03}. 168 of the 4039 verb types were marked as irregular.
We assigned verbs to train, development, and test sets according to a random 80-10-10 split. Each verb appears in exactly one of these sets once. This corresponds to a uniform distribution over types since every verb has an effective frequency of 1.
In contrast, the original \citeauthor{rumelhart1986learning} model was trained, and tested (data was not held out), on a set of 506 stem/past pairs derived from \cite{kuvcera1967computational}. 98 of the 506 verb types were marked as irregular.

\paragraph{Types vs Tokens.}
In real human communication, words follow a Zipfian distribution, with many
irregular verbs being exponentially more common than regular verbs. 
While this condition is more true to the external environment of
language learning, it may not accurately represent the psychological
reality of how that environment is processed. A body of
psycholinguistic evidence \cite{Bybee:95,Bybee:01,Pierrehumbert:01a} suggests
that human learners generalize phonological patterns based on the count of word {\em types} they appear in, ignoring the {\em token}
frequency of those types. Thus, we chose to weigh all verb types equally for training, effecting a uniform distribution over types as described above.

\paragraph{Hyperparameters and Other Details.}
Our architecture is nearly identical to that used in 
\cite{DBLP:journals/corr/BahdanauCB14}, with hyperparameters set following \citet[\S 4.1.1]{kann-schutze:2016:SIGMORPHON}.
Each input character has an
embedding size of 300 units. The encoder consists of a bidirectional
LSTM \cite{hochreiter1997long} with two layers. There is a dropout value of 0.3 between the
layers. 
The decoder is a unidirectional LSTM with two layers. Both the encoder
and decoder have 100 hidden units. Training was done using the
Adadelta procedure \cite{abs-1212-5701} with a learning rate of 1.0 and a minibatch size of 20.
We train for 100 epochs to ensure that all verb forms in the training
data are adequately learned.  We decode the model with beam search
($k=12$). The code for our experiments is derived from the OpenNMT package \cite{klein2017opennmt}.
We use accuracy as our metric of performance. We train the MGL as a non-neural baseline, using the code distributed with \cite{Albright:03} with default settings.

\paragraph{Results and Discussion.}
The non-neural MGL baseline unsurprisingly learns the regular past-tense pattern nearly perfectly, given that it is imbued with knowledge of phonological features as well as a list of phonologically illegal phoneme sequences to avoid in its output. 
However, 
in our testing of the MGL, the preferred past-tense output for all verbs was never an irregular formulation. This was true even for irregular verbs that were observed by the learner in the training set. One might say that the MGL is only intended to account for the regular route of a dual route system. However, the intended scope of the MGL seems to be wider. The model is billed as accurately learning `islands of subregularity' within the past tense system and Albright \& Hayes use the model to make predictions about which irregular forms of novel verb stems are preferable to human speakers (see discussion of wugs below).

\begin{table}
  \centering
  \setlength{\tabcolsep}{2.5pt}
   \begin{adjustbox}{width=.49\textwidth}
  {\small
  \begin{tabular}{l rrr rrr rrr} \toprule
    & \multicolumn{3}{c}{all} & \multicolumn{3}{c}{regular} & \multicolumn{3}{c}{irregular} \\ \cmidrule(lr){2-4} \cmidrule(lr){5-7} \cmidrule(lr){8-10}
                         & \multicolumn{1}{c}{train} & \multicolumn{1}{c}{dev} & \multicolumn{1}{c}{test} & \multicolumn{1}{c}{train} & \multicolumn{1}{c}{dev} & \multicolumn{1}{c}{test} & \multicolumn{1}{c}{train} & \multicolumn{1}{c}{dev} & \multicolumn{1}{c}{test} \\ \midrule
    Single-Task (MGL)    & 96.0    & 96.0  & 94.5  & 99.9    & 100.0 & 100.0  & 0.0    & 0.0   & 0.0         \\ \midrule
    Single-Task (Type)   & 99.8$\dagger$    & 97.4  & 95.1  & 99.9    & 99.2  & 98.9   & 97.6$\dagger$   & 53.3$\dagger$  & 28.6$\dagger$       \\
    Multi-Task (Type)    & 100.0$\dagger$   & 96.9  & 95.1  & 100.0   & 99.5  & 99.7   & 99.2$\dagger$   & 33.3$\dagger$  & 28.6$\dagger$        \\
    \bottomrule
  \end{tabular}
  }
 \end{adjustbox}
  \caption{Results on held-out data in English past tense prediction for single- and multi- task scenarios. The MGL achieves perfect accuracy on regular verbs, and 0 accuracy on irregular verbs. $\dagger$ indicates that a neural model's performance was found to be significantly different ($p < 0.05$) from the MGL.}
  \label{tab:results}
\end{table}

\begin{figure}
  \centering
    \begin{subfigure}{0.475\textwidth}
        \includegraphics[width=8.0cm]{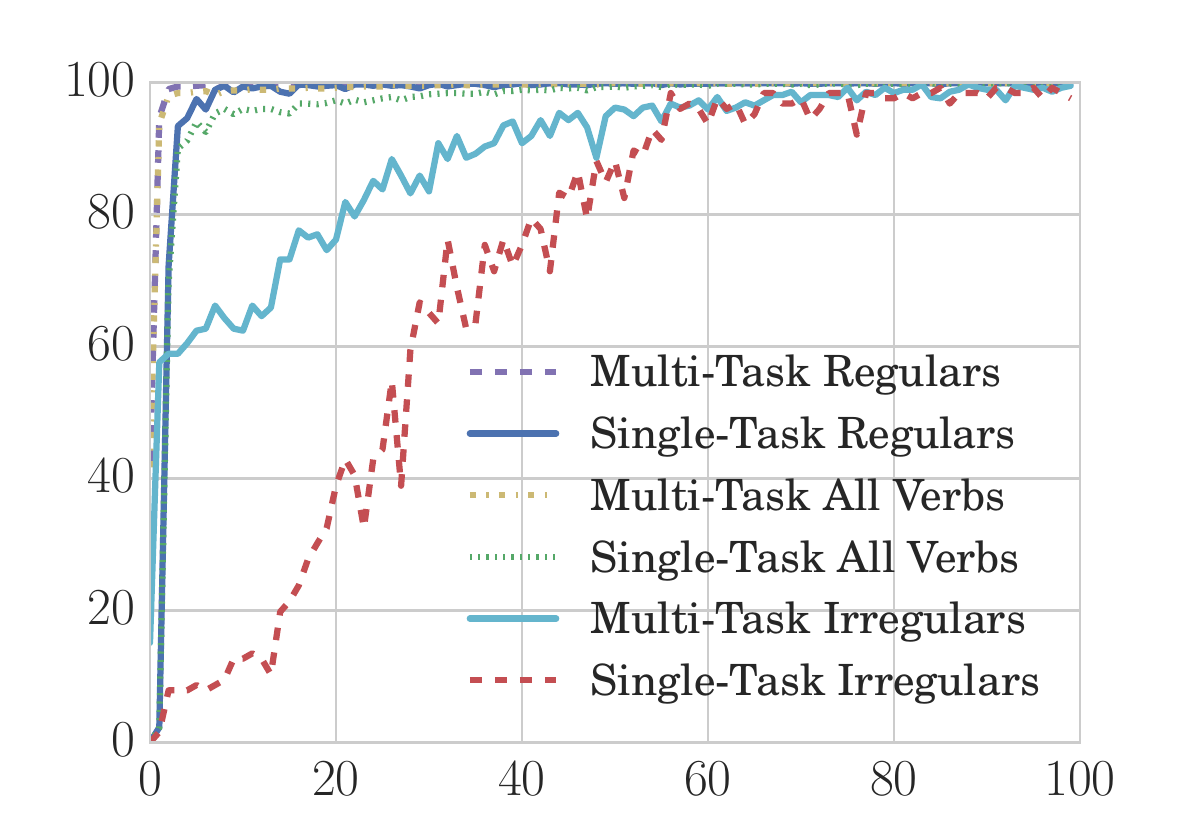}
    \end{subfigure}
    \vspace{.25cm}
    \caption{Single-task versus multi-task. Learning curves for the English past tense. The $x$-axis
      is the number of epochs (one complete pass over the training data) and the $y$-axis
      is the accuracy on the training data (not the metric of optimization).}
    \label{fig:curves}
\end{figure}

In contrast, the encoder--decoder model, despite no built-in knowledge of phonology, successfully learns to conjugate nearly all the verbs in
the training data, including irregulars---no reduction in scope is needed.  This
capacity to account for specific exceptions to the regular rule does not result in overfitting. We note
similarly high accuracy on held-out regular data---98.9\%-99.2\% at
convergence depending on the condition. We report the full accuracy in
all conditions in \cref{tab:results}. The $\dagger$ indicates when a neural model's performance was found to be significantly different ($p < 0.05$) from the MGL according to a $\chi^2$ test.  The encoder--decoder model achieves near-perfect accuracy on regular verbs, and irregular verbs seen during training, as well as substantial accuracy on irregular verbs in the dev and test sets. This behavior jointly results in better overall performance for the encoder--decoder model when all verbs are considered. \cref{fig:curves} shows learning
curves for regular and irregular verbs types in different conditions.

An error analysis of held-out data shows that the errors made by this
network do not show any of the problems of the \citeauthor{rumelhart1986learning}
architecture. There are no blend errors of the \word{eat}$\mapsto$\word{ated} variety. Indeed, the \emph{only} error the network makes on irregulars is 
overregularization (e.g., \word{throw}$\mapsto$\word{throwed}). In fact, the overregularization-caused lower
accuracy which we observe for irregular verbs in development and test, is expected
and desirable; it matches the human tendency to treat novel words as regular, lacking knowledge of irregularity
\cite{Albright:03}.

While most held-out irregulars are regularized, as expected, the encoder--decoder model \emph{does}, perhaps surprisingly, correctly conjugate a handful of irregular forms it has not seen during training---5 in the test set. However, three of these are prefixed versions of irregulars that exist in the training set (\word{retell}$\mapsto$\word{retold}, \word{partake}$\mapsto$\word{partook}, \word{withdraw}$\mapsto$\word{withdrew}). One (\word{sling}$\mapsto$\word{slung}) is an analogy to similar training words (\word{fling}, \word{cling}). The final conjugation, \word{forsake}$\mapsto$\word{forsook}, is an interesting combination, with the prefix \word{for}, but an unattested base form \word{sake} that is similar to \word{take}. \footnote{The sounds [s] and [t] are both coronal consonants, a fricative and a stop, respectively.}

From the training data, the only regular verb with an error is \word{compartmentalize}, whose past tense is predicted to be \word{compartmantalized}, with a spurious vowel change that would likely be ironed out with additional training. Among the regular verbs in the development and test sets, the errors also consisted of single vowel changes (the full set of these errors was \word{thin}$\mapsto$\word{thun}, \word{try}$\mapsto$\word{traud}, \word{institutionalize}$\mapsto$\word{instititionalized}, and \word{drawl}$\mapsto$\word{drooled}). 
Overall then, the encoder--decoder model performs extremely well, a far cry
from the $\approx$67\% accuracy of the \citeauthor{rumelhart1986learning} model. It exceeds any reasonable
standard of empirical adequacy, and shows human-like error
behavior.\looseness=-1

\begin{table}
\centering
\begin{tabular}{lrr} \toprule
English & \multicolumn{1}{c}{Network} & \multicolumn{1}{c}{MGL}\\ \midrule
Regular (\word{rife} $\sim$ \word{rifed}, n=58) & \textbf{0.48} & 0.35\\
Irregular (\word{rife} $\sim$ \word{rofe}, n=74) & \textbf{0.45} & 0.36\\
\bottomrule
\end{tabular}
  \caption{Spearman's $\rho$ of human wug production probabilities with MGL scores and encoder--decoder probability estimates.}
  \label{tab:wugcorrs}
\end{table}

\paragraph{Acquisition Patterns.}

\citeauthor{rumelhart1986learning} made several claims that their architecture 
modeled the detailed acquisition of the English past tense by children. The
core claim was that their model exhibited a \emph{macro} U-shaped
learning curve. Irregulars were
initially produced correctly, followed by a period of
overregularization preceding a final correct stage. However, \citeauthor{pinker1988language}
point out that \citeauthor{rumelhart1986learning} only achieve this pattern by manipulating the
input distribution fed into their network. They trained only on irregulars for a number of epochs, before flooding
the network with regular verb forms. \citeauthor{rumelhart1986learning} justify this by claiming that young
children's vocabulary consists disproportionately of irregular verbs early on,
but \citeauthor{pinker1988language} cite contrary evidence.
A survey of child-directed speech
shows that the ratio of regular to irregular verbs a child hears is
constant while they are learning their language \cite{Slobin:71}. Furthermore,
psycholinguistic results suggest that there is no early skew towards irregular verbs in the vocabulary children understand or use
\cite{Brown:73}.

\begin{table}
  \begin{adjustbox}{width=1.\columnwidth}
  \begin{tabular}{ll ll ll ll } \toprule
    \multicolumn{2}{c}{\sc cling} &  \multicolumn{2}{c}{\sc mislead} & \multicolumn{2}{c}{\sc catch} & \multicolumn{2}{c}{\sc fly} \\
    \# &  output & \# &  output & \# &  output & \# &  output \\ \cmidrule(r){1-2}  \cmidrule(r){3-4} \cmidrule(r){5-6} \cmidrule(r){7-8}
    5  & [kl\textipa{I}\textipa{N}d]  & 8  & [m\textipa{I}sl\textipa{i}\textipa{:}d\textipa{I}d]  & 7    & [k\textipa{\ae}t\textipa{S}]  & 6  & [fla\textipa{I}d] \\
    11 & [kl\textipa{2}\textipa{N}]   & 19 & [m\textipa{I}sl\textipa{E}{d}]                       & 31   & [k\textipa{\ae}t\textipa{S}]  & 31 & [flu\textipa{:}]  \\
    13 & [kl\textipa{I}\textipa{N}]   & 21 & [m\textipa{I}sl\textipa{E}{d}]                       & 43   & [k\textipa{O}t]               & 40 & [fla\textipa{I}d]  \\
    14 & [kl\textipa{I}\textipa{N}d]  & 23 & [m\textipa{I}sl\textipa{E}{d}]                       & 44   & [k\textipa{\ae}t\textipa{S}]  & 42 & [fl\textipa{e}\textipa{I}] \\
    18 & [kl\textipa{2}\textipa{N}]   & 24 & [m\textipa{I}sl\textipa{i}\textipa{:}d\textipa{I}d]  & 51   & [k\textipa{\ae}t\textipa{S}]  & 47 & [fla\textipa{I}d]   \\
    21 & [kl\textipa{I}\textipa{N}d]  & 29 & [m\textipa{I}sl\textipa{E}{d}]                       & 52   & [k\textipa{O}t]               & 56 & [flu\textipa{:}] \\
    28 & [kl\textipa{2}\textipa{N}]   & 30  & [m\textipa{I}sl\textipa{i}\textipa{:}d\textipa{I}d] & 66   & [k\textipa{\ae}t\textipa{S}]  & 62  & [fla\textipa{I}d]\\
    40 & [kl\textipa{2}\textipa{N}]   & 41  & [m\textipa{I}sl\textipa{E}d] & 73   & [k\textipa{O}t]               & 70  & [flu\textipa{:}] \\
    \bottomrule
  \end{tabular}
  \end{adjustbox}
  \caption{Here we evince the oscillating development of single words in our corpus. For each
    stem, e.g., {\sc cling}, we show the past form that was produced at change points to show
    the diversity of alternation. Beyond the last epoch displayed, each verb was produced correctly.
  }
  \label{tab:ushape}
\end{table}

While we do not wish to make a strong claim that the encoder--decoder architecture
accurately mirrors children's acquisition, only that it ultimately learns the correct generalizations, we wanted to see if it would display a child-like learning pattern \emph{without} 
changing the training inputs fed into the network over time, i.e., in all
of our experiments, the data sets remained fixed for all epochs, unlike
in \citeauthor{rumelhart1986learning}.
While we do not clearly see a macro U-shape, we \emph{do} observe \citepos{plunkett1991u} predicted oscillations for irregular learning---the so-called \emph{micro} U-shaped pattern. As shown in \cref{tab:ushape}, individual verbs oscillate between correct production and overregularization before they are fully mastered.\looseness=-1

\paragraph{Wug Testing.} As a further test of the MGL as a cognitive model, Albright \& Hayes created a set of 74 nonce English verb stems with varying levels of similarity to both regular and irregular verbs. For each stem (e.g., \word{rife}), they picked one regular output form (\word{rifed}), and one irregular output form (\word{rofe}). They used these stems and potential past-tense variants to perform a wug test with human participants. For each stem, they had 24 participants freely attempt to produce a past tense form. They then counted the percentage of participants that produced the pre-chosen regular and irregular forms (\textbf{production probability}). The production probabilities for each pre-chosen regular and irregular form could then be correlated with the predicted scores derived from the MGL. In  \cref{tab:wugcorrs}, we compare the correlations based on their model scores, with correlations comparing the human scores to the output probabilities given by an encoder--decoder model. As the wug data provided with \cite{Albright:03} uses a different phonetic transcription than the one we used above, we trained a separate encoder--decoder model for this comparison. Model architecture, training verbs, and hyperparameters remained the same. Only the transcription used to represent input and output strings was changed to match \cite{Albright:03}. Following the original paper, we correlate the probabilities for regular and irregular transformations separately. We apply Spearman's rank correlation, as we do not necessarily expect a linear relationship. We see that the encoder--decoder model probabilities are slightly more correlated than the MGL's scores.\looseness=-1

\subsection{Experiment 2: Joint Multi-Task Learning }

Another objection levied by \citeauthor{pinker1988language} is \citepos{rumelhart1986learning} focus on learning a
single morphological transduction: stem to past tense. Many
phonological patterns in a language, however, are not restricted to a
single transduction---they make up a core part of the phonological
system and take part in many different processes. For instance, the voicing assimilation patterns found in
the past tense also apply to the third person
singular: we see the affix \word{-s} rendered as [-s] after voiceless
consonants and [-z] after voiced consonants and vowels.

\citeauthor{pinker1988language} argue that the \citeauthor{rumelhart1986learning} model would not be able to take advantage of
these shared generalizations. Assuming a different network would need
to be trained for each transduction, e.g., stem to gerund and stem to
past participle, it would be impossible to learn that they have any
patterns in common. However, as discussed in \cref{sec:rnns}, a single encoder--decoder model can
learn multiple types of mapping, simply by tagging each input-output
pair in the training set with the transduction it represents. A
network trained in such a way shares the same weights and phoneme
embeddings across tasks, and thus has the capacity to
generalize patterns across all transductions, naturally capturing the
overall phonology of the language. Since different transductions
mutually constrain each other (e.g., English in general does not allow
sequences of identical vowels), we actually expect faster learning of
each individual pattern, which we test in the following
experiment.

We trained a model with an architecture identical to that used in
Exp. 1, but this time to jointly predict four mappings
associated with English verbs (past, gerund, past participle,
third-person singular).

\paragraph{Data.}
For each of the verb types in our base training set from Exp. 1, we
added the three remaining mappings. The gerund, past-participle, and
third-person singular forms were identified in CELEX according to
their labels in Wiktionary
\cite{sylakglassman-EtAl:2015:ACL-IJCNLP}. The network was trained on
all individual \emph{stem}$\mapsto$\emph{inflection} pairs in the
new training set, with each input string modified with additional
characters representing the current transduction \cite{kann-schutze:2016:SIGMORPHON}: \word{take {\tt
    $<${\footnotesize \textsf{PST}}$>$}}$\mapsto$\word{took}, but \word{take {\tt
    $<${\footnotesize \textsf{PTCP}}$>$}}$\mapsto$\word{taken}.\footnote{Without input annotation to mark the different mappings the network must learn, it would treat all input--output pairs as belonging to the same mapping, with each inflected form of a single stem as an equally likely output variant associated with that mapping. 
    Our network architecture cannot solve problems other than morphological transduction, e.g., discovering the range of morphological paradigm slots.}

\paragraph{Results.}
\cref{tab:results} and \cref{fig:curves} show the
results. Overall, accuracy is $>$99\% after convergence on
train. While the difference in final performance is never statistically significant compared to single-task learning, the learning curves are much steeper so this level of performance is achieved much more quickly. This provides evidence for our
intuition that cross-task generalization facilitates individual task
learning due to shared phonological patterning, i.e., jointly
generating the gerund hastens past tense learning.\looseness=-1

\section{Resolved and Outstanding Criticism}\label{sec:crit-summary}

In this paper, we have argued that the encoder--decoder architecture obviates many of the criticisms \citeauthor{pinker1988language} levied against \citeauthor{rumelhart1986learning}. Most importantly, the empirical performance of neural models is no longer an issue. The past tense transformation is learned nearly perfectly, compared to an approximate accuracy of 67\% for \citeauthor{rumelhart1986learning}. Furthermore, the encoder--decoder architecture solves the problem in a fully general setting. A single network can easily be trained on multiple mappings at once (and appears to generalize knowledge across them). No representational cludges such as Wickelphones are required---encoder--decoder networks can map arbitrary length strings to arbitrary length strings. This permits training and evaluating the encoder--decoder model on realistic data, including the ability to assign an exact probability to any arbitrary output string, rather than `representative' data designed to fit in a fixed-size neural architecture (e.g., fixed input and output templates). Evaluation shows that the encoder--decoder model does not appear to display any of the degenerate error-types \citeauthor{pinker1988language} note in the output of \citeauthor{rumelhart1986learning} (e.g., regular--irregular blends of the \word{ate}$\mapsto$\word{ated} variety).

Despite this litany of successes, some outstanding criticisms of \citeauthor{rumelhart1986learning} still remain to be addressed. On the trivial end, \citeauthor{pinker1988language} correctly point out that the \citeauthor{rumelhart1986learning} model does not handle homophones: \word{write}$\mapsto$\word{wrote}, but \word{right}$\mapsto$\word{righted}. This is
because it only takes the phonological make-up of the input string
into account, without concern for its lexical identity. This issue
affects the encoder--decoder models we discuss in this paper as well---lexical
disambiguation is outside of their intended scope. However, even the rule-learner
\citeauthor{pinker1988language} propose does not have such functionality.
Furthermore, if lexical markings {\em were} available, we could 
incorporate them into the model just as with different transductions in the multi-task set-up, i.e., by adding
the disambiguating markings to the input. 

More importantly, we need to limit any claims regarding treating encoder--decoder models as proxies for child language learners. \citeauthor{pinker1988language} criticized such claims from \citeauthor{rumelhart1986learning} because they manipulated the input data distribution given to their network over time to effect a U-shaped learning curve, despite no evidence that the manipulation reflected children's perception or production capabilities. We avoid this criticism in our experiments, keeping the input distribution constant. We even show that the encoder--decoder model captures at least one observed pattern of child language development---\citepos{plunkett1991u} predicted oscillations for irregular learning, the \emph{micro} U-shaped pattern. However, we did \emph{not} observe a macro U-shape, nor was the micro effect consistent across all irregular verbs. More study is needed to determine the ways in which encoder--decoder architectures do or do not reflect children's behavior. Even if nets do not match the development patterns of any individual, they may still be useful if they ultimately achieve a knowledge state that is comparable to that of an adult or, possibly, the aggregate usage statistics of a population of adults.

Along this vein, \citeauthor{pinker1988language} note that the \citeauthor{rumelhart1986learning} model is able to learn highly unnatural patterns
that do not exist in any language. For example, it is trivial to map
each Wickelphone to its reverse, effectively creating a mirror-image
of the input, e.g., \word{brag}$\mapsto$\word{garb}. Although an encoder--decoder model could likely learn linguistically unattested
patterns as well, some patterns may be more difficult to learn than
others, e.g., they might require increased time-to-convergence. It remains
an open question for future research to determine which patterns
RNN's prefer, and which changes are needed to account for over- and underfitting. Indeed, any sufficiently complex learning system (including rule-based learners) would have learning biases which require further study.

There are promising directions from which to approach this study. 
Neural networks are in a way analogous to
  animal models \cite{mccloskey1991networks}, in that they share interesting
  properties with human learners, as shown empirically, but are much easier and less costly
  to train and manipulate across multiple experiments. Initial experiments could focus on default architectures, as we do in this paper, effectively treating them as inductive baselines \cite{gildea1996learning} and measuring their performance given limited domain knowledge.
  We remark that encoder--decoder networks have no built-in knowledge of phonology or morphology.
Failures of these baselines would then point the way towards the biases required to learn human language, and models modified to incorporate these biases could be tested.

\section{Conclusion}
We have shown that the application of the encoder--decoder architecture
to the problem of learning the English past tense obviates many, though not all, of
the objections levied by \citeauthor{pinker1988language} against the first neural network proposed
for the task, suggesting that the criticisms do not extend to all
neural models, as \citeauthor{pinker1988language} imply.  Compared to a non-neural baseline, the encoder--decoder model accounts for both regular and irregular past tense formation in observed training data and generalizes to held-out verbs, all without built-in knowledge of phonology. While not necessarily intended to act as a proxy for a child learner, the encoder--decoder model also shows one of the development patterns that have been observed in children, namely a micro U-shaped (oscillating) learning curve for irregular verbs.
The accurate and substantially human-like
performance of the encoder--decoder model warrants consideration of its use as a research tool in
theoretical linguistics and cognitive science.

\bibliography{pinker-rebuttal}
\bibliographystyle{acl_natbib}

\end{document}